# Creating Intelligent Linking for Information Threading in Knowledge Networks

Dr T.R. Gopalakrishnan Nair
(Director): Research & Industry Incubation Center
Advanced AI and Bio Computing Research Group
D. S. Institutions
Bangalore, India
trgnair@yahoo.com, trgnair@ieee.org

Meenakshi Malhotra
(A. Member): Research & Industry Incubation Center
D. S. Institutions,
(Lecturer): Department of Information Science Engg
The Oxford College of Engineering,
Bangalore, India
uppal_meenakshi@yahoo.co.in

*Abstract* — **Informledge System (ILS) is a knowledge network with autonomous nodes and intelligent links that integrate and structure the pieces of knowledge. In this paper, we aim to put forward the link dynamics involved in intelligent processing of information in ILS. There has been advancement in knowledge management field which involve managing information in databases from a single domain. ILS works with information from multiple domains stored in distributed way in the autonomous nodes termed as Knowledge Network Node (KNN). Along with the concept under consideration, KNNs store the processed information linking concepts and processors leading to the appropriate processing of information.**

I. INTRODUCTION

The research in the field of knowledge processing has been active for the last few decades. Researchers in the field of knowledge representations and Ontology development has been concentrating on structuring the information.

There has been some development in the field of encoding domain knowledge in a fuzzy multi-layer perceptron (MLP) [1]. It restricts to domain knowledge being encoded but lacks in structuring of the knowledge network. Knowledge does not exist discretely from one domain alone but it becomes usually complete when the inter-domain knowledge is also linked. Although there have been efforts to interconnect heterogeneous knowledge bases [2], but these systems needed to be connected on the knowledge grid for doing so.

Conversely, there is no system at hand that would encode inter-domain knowledge effectively to provide a network of interlinked knowledge units that could be later used to retrieve information systematically and intelligently.

To overcome this, we had suggested Informledge System (ILS) in our work published earlier [3] [4]. In ILS, knowledge units belonging to same or different knowledge domains are linked together to form a knowledge network using the autonomous nodes and intelligent links. This interlinking of knowledge units for the knowledge network is by virtue of nodes with intelligence and multi-lateral links that have ingrained processing properties. These links play an important role in connecting correlated concepts.

This paper is organized as follows. Section II discusses Research Background. We present our Research Objective in section III. Section IV gives the Conceptual Model of the Informledge System and in section V we discuss the Link Dynamics of ILS and section VI details about system Simulation and Evaluation. Finally, we conclude in section VII.

II. RESEARCH BACKGROUND

The research in the field of knowledge management has shown that knowledge can be viewed in four different ways that primarily includes ontological, epistemological, commodity and community views of knowledge [5].

Knowledge bases have been constructed using ontologies. Ontology originated in the field of philosophy and is defined as a specification of what exists [6]. However in the field of computer science it is stated as "ontology is formal, explicit specification of a shared conceptualization". For different understanding of a concept stated in language, different representations are stated which involve ontological construction [7]. Standard Upper Ontologies [8] have been used to define general concepts at higher-level. However, it is not yet possible to represent the existing knowledge completely using one kind of ontology [9] [10].

Furthermore, ontology construction involves every concept to be represented in terms of words. Whereas in ILS, knowledge is represented by interlinked knowledge units, which are ordered set of entities [3]. This knowledge unit can represent a whole concept or a part of it [11]. The field of neuroscience also states that different regions of brain are responsible for concept formation and its representation into different languages. Every individual tend to think about a concept and then only represent that into a specific language [12].

Epistemological view is defined as the scientific view of knowledge. The commodity view understands knowledge as a static organizational resource, and community view assumes that knowledge is a social construct. Information stored in ILS can take up any of these forms by virtue of its homogeneous autonomous nodes and multilateral links.

Recently more focus is given on development of Brain Machine Interface (BMI) that aims at translating neuronal waves into a reality concept. BMI can translate raw neuronal signals into motor commands that reproduce arm reaching and hand grasping movements in artificial actuators [13]. It can be considered, if at all, it can be of any value to translate knowledge of brain to another format.

Researchers in the field of cognition and artificial intelligence have been endeavoring to simulate the knowledge of human brain. Although few have been able to understand the deep working patterns of neurons [16], but still a large amount of work need to be done to create knowledge bases that would be able to embed knowledge and links, and link each other in respective models. Our objective here is not to simulate the human brain per say but to map the information processing nature of the knowledge stored in the clustered neurons or any other nodes having ability to do processing. The knowledge units meaningfully connect to form the required knowledge as the clustered linked neurons connect to each other to produce a valid thought based on connected facts.

While creating knowledge bases emphasis has been on the word and sentences, which are part of information sharing, rather than the concepts [14]. But ILS emphasizes on inter-linking of concepts to encode and reproduce direct and derivative of them by pulling several threads through them.

Even human brain uses prior knowledge for decision making, through the use of concepts which are formed through abstraction, capture the shared meaning of similar entities [15]. In ILS decision making is backed by, subnormal manipulation of knowledge units.

### III. RESEARCH OBJECTIVE

Concepts linkages and the type-and-intensity of the linkage reveal the hidden structure. Together they explain the individual category in a coherent fashion. Through concepts, categories and relations that an individual make give understanding of our world. In ILS, we create knowledge dynamically from the existing knowledge units and links, while whatever was once embedded could only be extracted. Linkages form directed graph which can be an open or cyclic graph.

Knowledge unit is one entity along with its attributes. Creation of links leads to knowledge embedding. In ILS knowledge embedding is eventually connectivity of entities identified as autonomous nodes using ordered links. Meaningful links consist of coherent and incoherent knowledge. At present ILS embeds only coherent knowledge, which is explicitly embedded and does not include the knowledge formed by cross-linking of knowledge units. The knowledge of ILS is like a sphere as shown in Fig. 1 whose inner-most sphere consists of the interlinked knowledge units as part of a knowledge, outer to it is the knowledge encoder (KE) [2] which helps in knowledge embedding and retrieval, next to it is the knowledge validation sphere which helps in validation of this knowledge with the stored facts in ILS and the outer-most sphere provide the knowledge representation to interlinked knowledge units to form an understandable knowledge.

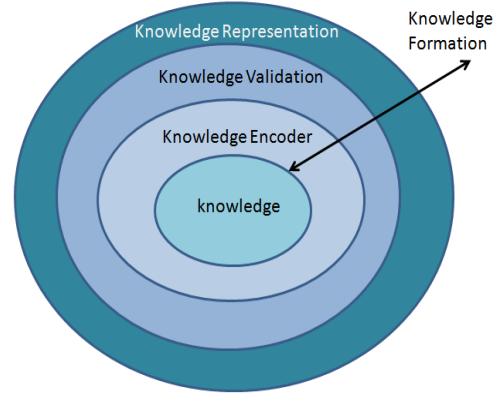

Figure 1. Sphere of Knowledge

In our previous paper, we had talked about the knowledge embedding and retrieval strategies [3]. Links which form the vital part of ILS helps in linking knowledge units intelligently. The intelligence of the system lies with linkages. In the following sections we discuss about ILS and how the link decides where to connect.

### IV. INFORMLEDGE SYSTEM

ILS is a modified knowledge network system dealing with logical storage and connectivity of information base to form knowledge using autonomous nodes and multi-lateral links [3]. The node in ILS is termed as Knowledge Network Node (KNN). KNNs are autonomous by virtue of their capability to store, infer and reason the creation of the knowledge thread. The nodes in the ILS are homogeneous. The structure of the KNN encompasses four quadrants namely input/output, storage and parser, link database and link manager, as discussed in our last paper [2].

Links are the significant constituent of ILS. They characterize the intelligence of the system. During knowledge embedding the knowledge units, KNNs, get connected based on the following link properties: directional and performance properties, which include exclusive or inclusive, additive or subtractive and integrative or differentiative [3]. In the following section we discuss how these properties help in building the knowledge thread.

### V. LINK DYNAMICS

ILS is an organized system in which connectivity and search is done through an ordered set of links. Properties of links leads to processing where by several logics can be retrieved through it. Intelligence of understanding is in links which includes classifying, correlating and extrapolating the information.

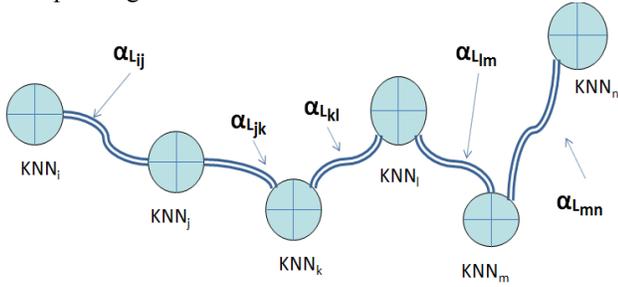

Figure 2. Knowledge Thread formed by linking KNNs

Knowledge thread is like a string of beads where the knowledge units, KNN, are the beads and the thread connecting the knowledge units is the link. The knowledge thread shown in Fig. 2 can be represented as:

Knowledge Thread, $K_t = \alpha_{Lij} + \alpha_{Ljk} + \alpha_{Lkl} + \alpha_{Llm} + \alpha_{Lmn}$

$$= \sum_{x=i,y=j}^{m,n} (\alpha_{Lx,y}) \quad (1)$$

In Equation (1), αLij represent the link between the i$^{th}$ and j$^{th}$ nodes that is KNNi and KNNj. From equation (1), it is clear that the knowledge thread formed after linking n nodes is summation of all the links. $\alpha_{L12}$ is defined as combination of all the three properties between the two nodes as follows:

$$\alpha_{L12} = (\alpha_{p1 1,2} + \alpha_{p2 1,2} + \alpha_{p3 1,2}) \quad (2)$$

Hence, equation (1) could now be expanded using equation (2) to

$$K_t = \sum_{x=i,y=j}^{m,n} \left( (\alpha_{p1x,y} + \alpha_{p2x,y} + \alpha_{p3x,y}) + \cdots \right)$$

Thus for any link between KNNx and KNNy, the link formed is the combination of the properties p1, p2, p3 and p4, As discussed in our paper[3], p1 denotes the directional properties(source/destination KNN), p2 is a combination of three performance properties and p3 is the temporal property. The performance property, p2 gets a value which is a combination of the link properties that includes the following:

- P21 – Additive / Subtractive
- P22 – Inclusive / Exclusive
- P23 – Integrative / Differentiative

A. Structural Rules

While substituting the values for the structural properties of a link the following structural rules are followed:

*1) And/or connectivity between concepts is not listed explicitly:* To represent AND the two concepts are individually linked to the concerned fact. The same is represented logically as shown in Fig. 3. The Fig. 3 shows that if the knowledge is "Computer can store, retrieve and process information" then the three concepts about storage, retrieval and processing are linked to the computer concept incoherently.

*2) Each knowledge unit is a concept:* Each knowledge unit is like a concept and each concept can be represented via a KNN i.e. each concept is a KNN.

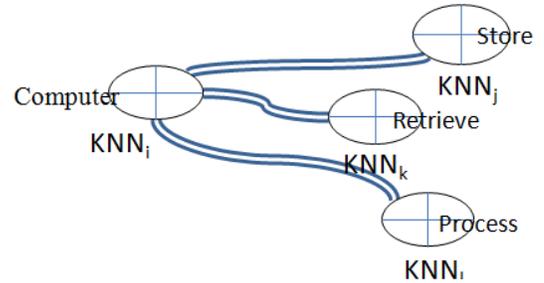

Figure 3. Three links showing the AND connection

*3) Assigning the performance properties*

*a) Additive/Subtractive:* Some of the concepts add some additional information to other concept, in those cases the link connecting those concepts are called additive. Consider the knowledge "Some plants have big leaves like banana", here the two concepts represented by KNN (big) and KNN (leaves) share the additive property. Negation between two concepts is represented by subtractive property e.g. for the knowledge "Some plants cannot stand straight", so the two concepts KNN (stand) and KNN (straight) have the link with subtractive property.

*b) Integrative/Differentiative:* One concept is said to share integrative property with other concepts when they aggregate to form a single concept eg considering the knowledge "computer have one CPU and main memory", CPU and main memory concepts integrate to form a computer concept.

*c) Inclusive/Exclusive:* The two concepts that are linked via inclusive property share the 'belongs-to' relationship. If that relationship does not exist then the link is marked to have exclusive relationship. Eg: botany is a branch of biology- has two main concepts botany and biology whereas branch linking attribute depicts that botany is inclusive of biology.

## B. Link Formation

Knowledge embedding is micro-fined using millions of knowledge nodes via highly processing intelligent links. List of links are stored by the parser in Link Database and the parser takes the direction to complete the knowledge thread. Link Manager comprises of the processor channel which process the directional and the performance properties of the links. The link is a combination of directional and the weightage of the three performance properties. The link may have value for one, two or all the three of these performance properties as shown in Fig. 4.

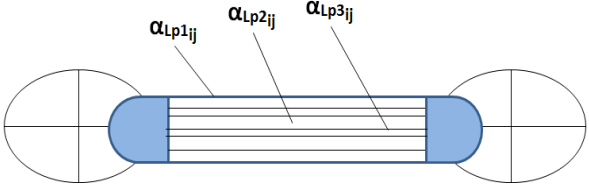

Figure 4. Link Properties

There are natural links which are created during knowledge embedding. In addition to that there are unnatural links that are the result of interconnectivity of the links. Now as the knowledge is retrieved the various knowledge components, KNNs, which were earlier not connected are also linked which would span to multiple planes on which the components exists as shown in Fig. 5.

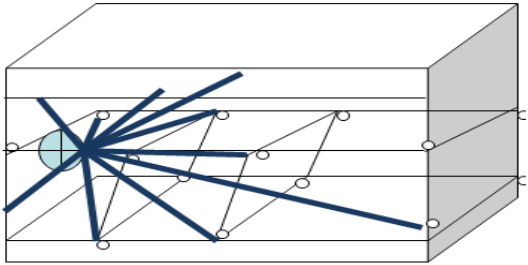

Figure 5. KNN linked to other KNNs on multiple planes

These unnatural links are validated during knowledge retrieval when the knowledge pass through the knowledge validation sphere.

## VI. SIMULATION AND EVALUATION

The link size grows as the knowledge grows. As more and more knowledge is embedded into the system the knowledge thread increases. In our simulation, we embedded knowledge from different domains as what exist in the world, Continent, Computers, etc.

ILS is synonymous to a box of beads from which if we pull out a bead different thread of beads could be taken out. Similarly, from ILS if we take a concept, we can retrieve different knowledge threads from it. Every knowledge thread extracted would have different number of links processed.

TABLE 1  THREAD LINKS FOR KNNS

| KNN | Number of Links in each Thread | | | | | | | | | |
|---|---|---|---|---|---|---|---|---|---|---|
| Knn11 | 1 | 2 | 7 | 3 | 4 | 4 | 7 | 10 | 11 | 8 | 9 |
| Knn72 | 3 | 4 | 4 | 7 | 7 | 7 | 7 | 7 | | | |
| Knn16 | 2 | 2 | 5 | 6 | 6 | 10 | 10 | 10 | 10 | 10 | 10 |

Table 1 depicts the simulation result of retrieving knowledge threads for three KNNs from the pool of linked knowledge nodes. The strength of the thread, which is given by number of links processed in a thread, varies in the three cases. For knowledge node, knn11 few threads span to second and fourth level while few threads went up to eleventh level.

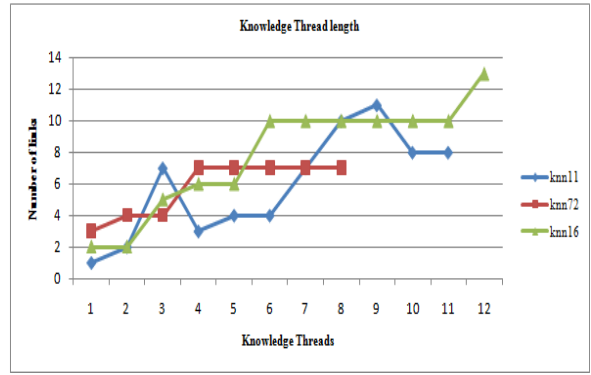

Figure 6. Knowledge Thread Lengths

The same has been depicted in Fig. 6. The gridlines in the figure denote the links connected for each thread. In case of knn11, the first knowledge thread has only one link, third one has seven and the ninth thread has eleven links. As more and more knowledge is embedded, the length of the thread retrieved also increases.

Table 2 depicts that the number of threads retrieved from a knowledge node would be more if the KNN at higher level of knowledge cone whereas it would be quite less for the KNN which is at the base of the knowledge cone.

TABLE 2 THREADS RETRIEVED

| Knowledge Node | knn16 | knn39 | knn53 | knn74 | knn96 |
|---|---|---|---|---|---|
| Number of Threads Retrieved | 28 | 12 | 18 | 10 | 1 |

## VII. CONCLUSION

In this paper, we presented the Informledge approach for creating information by link dynamics which intelligently process the information to form a network of knowledge. The link processor at the links formalizes an ordered collection of links by processing the performance properties. ILS has been validated to work for few domains, leading to formation of small and light-weight links. The future work include formulating links that traverse across multiple domains leading to bigger and heavy-weight links. This also includes the heuristic search involved in the link selection during knowledge embedding and retrieval. The future work also includes validation of the retrieved knowledge across the standard logic of assertions.

REFERENCES

[1] Mohua Banerjee, Sushmita Mitra, and Sankar K. Pal, Rough Fuzzy MLP: Knowledge Encoding and Classification IEEE TRANSACTIONS ON NEURAL NETWORKS, VOL. 9, NO. 6, NOVEMBER 1998

[2] Takafumi Nakanishi, Koji Zettsu, Yutaka Kidawara, Yasushi Kiyoki, Approaching the Interconnection of Heterogeneous Knowledge Bases on a Knowledge Grid, Fourth International Conference on Semantics, Knowledge and Grid, IEEE 2008

[3] T.R. Gopalakrishnan Nair, Meenakshi Malhotra, Informledge System: A modified Knowledge Network with Autonomous Nodes using Multi-lateral Links, International Conference, International Conference on Knowledge Engineering and Ontology Development, KEOD 2010, Proceeding of KEOD-2010, pp 351-354, Valencia-Spain, October 2010.

[4] T.R. Gopalakrishnan Nair, Meenakshi Malhotra, Informledge System: Knowledge Embedding and Retrieval Strategies in an Informledge System

[5] Maria Jakubik, Exploring the knowledge landscape: four emerging views of knowledge, JOURNAL OF KNOWLEDGE MANAGEMENT VOL. 11 NO. 4 2007, pp. 6-19,

[6] Thomas R. Gruber, 1993. A translation approach to portable ontology specifications. In: Knowledge Acquisition. 5: 199-220.

[7] Christopher Brewster, Kieron O'Hara, Knowledge Representation with Ontologies: The Present and Future, 2004 IEEE Intelligent Systems.

[8] Ian Niles, Adam Pease, 2001. Towards a Standard Upper Ontology. 2nd International Conference FOIS-2001.

[9] Adam Pease,2010. Ontology Development Pitfalls. Retrieved October 3, 2005, from http://www.ontologyportal.org /Pitfalls.htm

[10] H´ector D´ıez-Rodr´ıguez and Guillermo Morales-Luna, Jos´e Oscar Olmedo-Aguirre,2008. Ontology-based Knowledge Retrieval. Seventh Mexican International Conference on Artificial Intelligence

[11] Doan A, Madhavan J, Dhamankar R, et al. learning toMatch Ontologies on the Semantic Web. In VLDB Journal,Special Issue on the Semantic Web, November 2003.Vol12(4), pp. 303-319.

[12] Eric R. Kandel, James H. Schwartz, Thomas M. Jessel: "Principles of Neural Science", ISBN-0-07-112000-9

[13] Mikhail A. Lebedev and Miguel A.L. Nicolelis, "Brain–machine interfaces: past, present and future", TRENDS in Neurosciences Vol.29 No.9

[14] Rajendra Akerkar,Preeti Sajja: "Knowledge-Based Systems", ISBN-13:978076377647

[15] Dharshan Kumaran, Jennifier J. Summerfield, Demis Hassabi, Eleanor A. Maguire: Tracking the Emergence of Conceptual Knowledge during Human Decision Making, Neuron Volume 63, Issue 63, 889-901, September 2009, Elsevier Inc.

[16] Moritz Helmstaedter,1 Kevin L Briggman1, Winfried Denk1 , High-accuracy neurite reconstruction for high-throughput neuroanatomy, Nature Neuroscience; Volume: 14; Pages: 1081–1088; Year published: (2011)